\documentclass[runningheads]{llncs}
\usepackage{graphicx}
\usepackage{multirow}
\usepackage{url}
\usepackage{subfig}
\usepackage{amsmath}
\usepackage{siunitx}
\usepackage{booktabs}
\usepackage{subscript}

\usepackage{tabularx}
\usepackage{colortbl}
\usepackage{hhline}

\usepackage[multiple]{footmisc}

\begin{document}
\title{Recursive Style Breach Detection\\ with Multifaceted Ensemble Learning}
%
%
\newcommand*\samethanks[1][\value{footnote}]{\footnotemark[#1]}

\author{
Daniel Kopev\thanks{Equal Contribution}\inst{1} \and
Dimitrina Zlatkova\samethanks[1]\inst{1} \and
Kristiyan Mitov\samethanks[1]\inst{1} \and
Atanas Atanasov\samethanks[1]\inst{1} \and
Momchil Hardalov\inst{1}\and
Ivan Koychev\inst{1}\and
Preslav Nakov\inst{2}
}
\authorrunning{D. Kopev, D. Zlatkova, K. Mitov, A. Atanasov et al.}
%
\institute{FMI, Sofia University ``St. Kliment Ohridski'', Sofia, Bulgaria\\
\email{\{dkopev, dvzlatkova, kmmitov, amitkov\}@uni-sofia.bg}\\
\email{\{hardalov, koychev\}@fmi.uni-sofia.bg} 
\and
Qatar Computing Research Institute, HBKU, Doha, Qatar\\
\email{pnakov@qf.org.qa}}
\maketitle              
\begin{abstract}
We present a supervised approach for style change detection, which aims at predicting whether there are changes in the style in a given text document, as well as at finding the exact positions where such changes occur. In particular, we combine a TF.IDF representation of the document with features specifically engineered for the task, and we make predictions via an ensemble of diverse classifiers including SVM, Random Forest, AdaBoost, MLP, and LightGBM. Whenever the model detects that style change is present, we apply it recursively, looking to find the specific positions of the change.
Our approach powered the winning system for the PAN@CLEF 2018 task on Style Change Detection.


\end{abstract}

\section{Introduction}

There are numerous tasks related to authorship attribution, but most of the research has been concentrated on large documents. An interesting problem to tackle for smaller texts is the one of style change detection: given a text document, identify whether style change occurs anywhere in it. This formulation usually entails a uniform distribution of text segments from multiple authors. A version of it is the intrinsic plagiarism detection problem, in which it is considered that there is a dominant author of the document being examined. Another variation is the task of detecting style change positions: determine the exact location of style breaches in the text. Historically, this has proven to be a difficult but interesting task, and performance in terms of accuracy has been low, leaving a lot of room for potential improvements over the state-of-the-art. Here, we target two tasks: (\emph{i})~predicting whether style change occurs (Style Change Detection\footnote{\url{http://pan.webis.de/clef18/pan18-web/author-identification.html}}), and (\emph{ii})~finding the exact position of the change (Style Breach Detection\footnote{\url{http://pan.webis.de/clef17/pan17-web/author-identification.html}}).

\section{Related Work}

\paragraph{Authorship attribution} Previous work on authorship attribution and related problems (e.g., author obfuscation \cite{stein:2017n,CLEF2017:SU,PAN2016:SU,stein:2016k}) used primarily term frequencies~\cite{karas:2017,kuznetsov:2016} and features from stylometry~\cite{khan:2017a,kuznetsov:2016,sittar:2016}. We borrowed ideas for traditional features from~\cite{10.1007/978-3-540-70981-7_40,pervaz:2015}, but we also designed some new ones, related to tautology, grammar contractions, quote use discrepancies, beginning and ending author statement words, and named entity spellings (see Section~\ref{features}).

\paragraph{Style breach detection.}
See \cite{stein:2017m} for a summary of previous work on style breach detection and related tasks. 
Here we outline some of the most relevant work.
Kara{\'s} et al.~\cite{karas:2017} used TF.IDF, POS tags, stop words and punctuation to represent paragraphs in the text, and applied a Wilcoxon Signed Rank test to check whether two segments are likely to come from the same distribution. They moved a sliding window over the sentences, computing similarity statistics and using dictionaries with common English words and sentiment. Then, they used a predefined threshold to determine whether a style breach between two sentences was likely. Safin and Kuznetsova~\cite{safin:2017} explored techniques typically used for intrinsic plagiarism detection. They vectorized sentences using pre-trained skip-thought models and looked for outliers using cosine-based distance between vectors.

\section{Style Change Detection}
\label{style_change_detection}

Here, we describe our approach, which powered the winning system \cite{CLEF2018:PAN:StyleBreachDetection:FMI} for the PAN@CLEF 2018 task on Style Change Detection \cite{CLEF2018:PAN:StyleBreachDetection}.


\subsection{Data}

We used data provided by the organizers of the CLEF-2018 PAN task on Style Change Detection\footnote{\url{http://pan.webis.de/clef18/pan18-web/author-identification.html}} \cite{CLEF2018:PAN:StyleBreachDetection}, which was based on user posts from StackExchange covering different topics with 300--1,000 tokens per document. It included a training set of 3,000 documents and a validation set of 1,500 documents.

\subsection{Preprocessing}
\label{preprocessing}

We pre-process the data in two phases. The first phase is applied before any feature extraction has taken place, and it replaces URLs and long numbers with specific tokens. The second phase is applied during feature computation. It filters the stream of words and replaces file paths, long character sequences and very long words with special tokens. Additionally, an attempt is made to split long hyphenated words (with three or more parts) by checking whether most of the sub-words are present in a dictionary of common words (from the NLTK words corpus~\cite{Loper:2002:NNL:1118108.1118117}). The objective of all these steps is to reduce the impact of long words, which could adversely affect features that take word length into account. Such features are those from the lexical group and preprocessing is applied to them only, since it might have undesirable effect on the rest of the features.

\subsection{Text Segmentation}

\begin{figure}
\includegraphics[width=12cm,keepaspectratio]{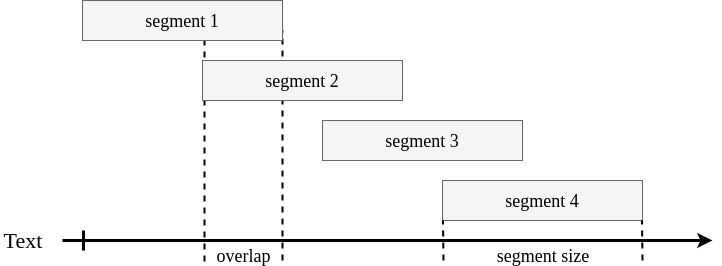}
\caption{Applying a sliding window on the text.}
\label{fig:sliding_window}
\end{figure}

Style changes in text documents entail that parts of the text would differ in some way. In an attempt to spot such differences, we split the document into four segments of roughly equal length (measured in terms of word tokens), we calculated the feature vectors for each of the segments, and we found the maximum difference between the values for each feature for any pair of segments. We chose the number of segments (namely, four) based on the distribution of the number of style changes across the entire training dataset. In order to obtain more potential data points, we applied a sliding window across each document with an overlap of one third of the segment size (see Figure~\ref{fig:sliding_window}). We applied this segmentation procedure for four of the feature groups, three of which used a sliding window. See Section~\ref{features} for more details.

\subsection{Text Representation}
\label{features}

Below, we describe the features we engineered specifically for the task of discovering style changes. The dimensionality of each of them is shown in Table~\ref{feature_dimensionality}.

\begin{table}
\centering
\begin{tabular}{lc}
\toprule
\textbf{Features} & \textbf{Dimensionality}\\
\midrule
Tautology & 5\\
Grammar contractions & 29\\
Beginning and ending of author statements & 1\\
Quotation marks & 1\\
Readability & 9\\
Frequent words & 415\\
Lexical & 34\\
Vocabulary richness & 2\\
Named entity spellings & 2\\
\bottomrule
\end{tabular}
\caption{Dimensionality of our features.}\label{feature_dimensionality}
\end{table}

\textbf{Tautology:}
At a grammatical and, one might say, psychological level, writers attempt to avoid using repetitive statements. We account for this by looking at the average number of occurrences of each one to five word-grams in the entire document, and we use a vector of size five with the respective averages for text representation.

\textbf{Grammar Contractions:}
Another viewpoint we look at is based on the discrepancies in words, which have contracted forms or shortened version of words such as {\it I will (I'll)}, {\it are not (aren't)}, and {\it they are (they're)}. Because most people favor one or the other, contraction apostrophes are suitable discriminative features (even more so in formal contexts) for identifying whether a piece of text is likely to be single- or multi-authored.

\textbf{Frequent Words:}
Frequent words include stop words (taken from NLTK~\cite{Loper:2002:NNL:1118108.1118117}) and function words (compiled from three separate lists\footnote{\url{http://semanticsimilarity.files.wordpress.com/2013/08/jim-oshea-fwlist-277.pdf}}\footnote{\url{http://www.sequencepublishing.com/1/academic.html}}\footnote{\url{http://www.edu.uwo.ca/faculty-profiles/docs/other/webb/essential-word-list.pdf}}). Each frequent word is counted per text segment.

\textbf{Lexical:}
The lexical features are computed as ratios per text segment using a sliding window and can be divided into the following three types:

\textit{Character-based:}
Spaces, digits, commas, colons, semicolons, apostrophes, quotes, parenthesis, number of paragraphs, and punctuation in general.

\textit{Word-based:}
We POS-tag the segment using NLTK, and we extract features such as ratios of pronouns, prepositions, coordinating conjunctions, adjectives, adverbs, determiners, interjections, modals, nouns, personal pronouns and verbs. Other word-based features include words with 2 or 3 characters, words with over 6 characters, average word length, all-caps words, and capitalized words.

\textit{Sentence-based:}
Those include ratios of question, period, exclamation sentences, short and long sentences, and average sentence length.

\textbf{Quotation Marks:}
Normally used in pairs, different people might consistently prefer using either single or double quotation marks. We first exclude every shortened word with apostrophe (from a grammar contraction words dictionary), and then we use the variance in the number of single and double quotes as a single-feature representation of the documents.

\textbf{Vocabulary Richness:}
Similarly to~\cite{10.1007/978-3-540-70981-7_40}, vocabulary richness is represented by averaged word frequency class. Using the Google Books common words list,\footnote{\url{http://norvig.com/google-books-common-words.txt}} the frequency class of a word $x$ is computed as $\log_2{\frac{f(X)}{f(x)}}$, where $f$ is the frequency function and $X$ is the most frequent word in the corpus, in our case \textit{the}. Two features are extracted per segment: the average frequency class of all words in it, and the ratio of unknown words (words not present in the common words list).

\begin{table}[tbh]
\centering
\subfloat[beginning]{
\begin{tabular}{p{2cm}c}
\toprule
\textbf{Word} & \textbf{Rescaled Counts}\\
\midrule
however & $\num[round-mode = places]{1.0}$ \\
one & $\num[round-mode = places]{0.9615384615384616}$ \\
note & $\num[round-mode = places]{0.9230769230769231}$ \\
edit & $\num[round-mode = places]{0.8653846153846154}$ \\
first & $\num[round-mode = places]{0.6923076923076923}$ \\
yes & $\num[round-mode = places]{0.6346153846153846}$ \\
also & $\num[round-mode = places]{0.5961538461538461}$ \\
another & $\num[round-mode = places]{0.5}$ \\
finally & $\num[round-mode = places]{0.40384615384615385}$ \\
since & $\num[round-mode = places]{0.36538461538461536}$ \\
update & $\num[round-mode = places]{0.34615384615384615}$ \\
let & $\num[round-mode = places]{0.3076923076923077}$ \\
\bottomrule
\end{tabular}
}
\qquad
\subfloat[ending]{
\begin{tabular}{p{2cm}c}
\toprule
\textbf{Word} & \textbf{Rescaled Counts}\\
\midrule
etc & $\num[round-mode = places]{1.0}$ \\ 
time & $\num[round-mode = places]{0.9473684210526315}$ \\ 
well & $\num[round-mode = places]{0.7631578947368421}$ \\ 
question & $\num[round-mode = places]{0.6842105263157895}$ \\ 
way & $\num[round-mode = places]{0.6052631578947368}$ \\ 
god & $\num[round-mode = places]{0.5789473684210527}$ \\ 
p & $\num[round-mode = places]{0.4473684210526316}$ \\ 
example & $\num[round-mode = places]{0.4473684210526316}$ \\ 
work & $\num[round-mode = places]{0.39473684210526316}$ \\ 
war & $\num[round-mode = places]{0.39473684210526316}$ \\ 
though & $\num[round-mode = places]{0.3684210526315789}$ \\ 
answer & $\num[round-mode = places]{0.3684210526315789}$ \\ 
\bottomrule
\end{tabular}
}
\caption{The 12 most frequent beginning and ending words in author statements (after stopword removal).}
\label{splitpoints}
\end{table}

\textbf{Readability:}
The following readability features are computed per text segment via the \textit{Textstat}\footnote{\url{http://github.com/shivam5992/textstat}} Python package: Flesch reading ease, SMOG grade, Flesch-Kincaid grade, Coleman-Liau index, automated readability index, Dale-Chall readability score, difficult words, Linsear write formula, and Gunning fog.

\textbf{Beginning and Ending of Author Statements:}
As can be seen in Table~\ref{splitpoints}, author statements begin and end with very different types of words. This can be used to locate points in documents where word clusters of small size contain high amount of these terms. We tried two approaches, applied after stopword removal, and we experimented with word phrases of sizes 1, 2 and 3, with single-terms yielding the best results. The first approach assigns scores to words to be rescaled (min-max normalized): it counts the number of times the target word is at a beginning or at an ending position relative to the author statement. A bit more sophisticated approach scores words based on how close they are to such a position. Each word is processed using a very steep half-sigmoid function (Equation~\ref{sigmoid_1}, with \textit{k} denoting the steepness), taking its relative position and rewarding those that are extremely close to a beginning or to an ending. Then, each word list of position scores is averaged across all documents.
\begin{equation}
\label{sigmoid_1}
\begin{aligned}
x = \dfrac{\lvert\dfrac{statementLength}{2} - (position + 1)\rvert}{\dfrac{statementLength}{2}}\\
Score(position\textsubscript{statement}) = \dfrac{(0 + k) * x}{(1 + k) - x}
\end{aligned}
\end{equation}

\noindent Finally, the document feature vector is represented by looking at local document clusters of three words, containing multiple high-scored words, indicating there may be an end of author statement immediately followed by a beginning of a new one. This document representation was not added as part of the stacking classifier (Section~\ref{stacking}), but nevertheless has a strong performance on its own, yielding 65\% accuracy.

\textbf{Named Entity Spellings:}
Different named entity spellings can reflect personal preferences, rather than cultural ones. We use Damerau-Levenshtein string-edit distance~\cite{Damerau:1964:TCD:363958.363994,1966SPhD...10..707L} to find inconsistencies in the wording of the same named entities within an edit distance of 1. The feature vector consists of the minimum counts between the different spellings for each found named entity.

\subsection{Classification}

\subsubsection{LightGBM}
\label{lightgbm}

Our gradient boosting approach combines LightGBM \cite{NIPS2017_6907} with Logistic Regression and TF.IDF vector representation. 
Note that we use the test data when calculating the IDF statistics.
This is not cheating as we do not use the labels for the examples, we only calculate word frequencies. Then, we use feature importance weights with a Logistic Regression estimator to select the best TF.IDF features;
moreover, we only select features with weight greater than 0.1.
We tune the Logistic Regression hyper-parameters using cross-validation. The best results are achieved with Stochastic Average Gradient descent, and inverse of the regularization strength C of 2. 
We trained using bagging with five folds. A simple LightGBM baseline achieved 73\% accuracy on the validation set. Tuning the LightGBM hyper-parameters increased the accuracy to 86\%, supported by a CV score of 85\%. These parameters can be seen in Table~\ref{table_stacking}. 

\subsubsection{Stacking}
\label{stacking}

\begin{figure}[tbh]
\includegraphics[width=12cm,keepaspectratio]{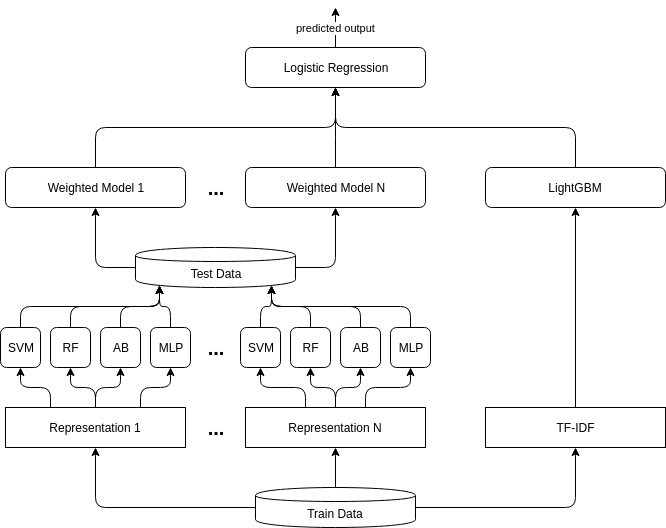}
\caption{Training the stacking classifier.}
\label{fig:stacking_train}
\end{figure}

The basic idea behind our Stacking Ensemble classifier was to take into account different independent points of view in the context of distinguishing multi-authored documents and to learn dependencies between them. At the bottom level, we train four different non-linear classifiers --- SVM, random forest, AdaBoost trees, multi-layer perceptron (described in Table~\ref{table_stacking}) --- for each feature vector derived from the representations in Section~\ref{features} on 75\% of the training data. Then, each one of them predicts on the remaining 25\% and is assigned a weight, based on its accuracy, relative to the remaining classifiers with the same input feature vectors. Those groups form a single vector each with prediction class probabilities, based on the weights and the outputs of its classifiers. These vectors, together with the predictions of the LightGBM classifier (see Section~\ref{lightgbm}), serve as an input to a simple linear Logistic Regression meta-learner. The process of training is visualized in Figure~\ref{fig:stacking_train}.

Before predicting, each classifier is trained again on the whole dataset (except for LightGBM, which is not weighted across groups). For each new sample, the zero-level classifiers use the same weights learned in training to transform the given sample as an input vector of probabilities for the meta-learner.
This yields accuracy of 87\%.
The coefficients learned by the meta-model for each text representation and their standalone accuracies can be seen in Table~\ref{feature_weights}.

\begin{table}[tbh]
\centering
\begin{tabular}{lcc}
\toprule
\textbf{Representation} & \textbf{Coefficient} & \textbf{Accuracy}\\
\midrule
Tautology & 1.50 & 67.4\\
Grammar Contractions & 1.25 & 61.0\\
Quotation Marks & 0.05 & 55.8\\
Readability & -0.25 & 61.0\\
Frequent Words & 0.81 & 63.3\\
Lexical & 0.27 & 64.9\\
Vocabulary Richness & -1.46 & 51.0\\
LightGBM with TF-IDF & 5.01 & 88.5\\
\bottomrule
\end{tabular}
\caption{Style Change Detection: model coefficients and accuracy for different feature representations (in isolation).}\label{feature_weights}
\end{table}

\begin{table}[tbh]
\begin{tabular}{cll}
\toprule
\textbf{Classifier} & \textbf{Hyper-parameter} & \textbf{Value}\\
\midrule
\multirow{4}{*}{Support Vector Machine} & kernel & Radial basis function\\
& penalty $C$ & 1.0\\
& tolerance & 0.001\\ 
& class weight & balanced\\
\midrule
\multirow{2}{*}{Random Forest} &  estimators & 300\\
& with replacement & Yes\\
\midrule
\multirow{2}{*}{AdaBoost Trees} & base estimator & Decision tree\\
& estimators & 300\\
\midrule
\multirow{9}{*}{Multi-layer Perceptron} & layers & 1\\
& layer size & 100\\
& activation & ReLU\\
& optimization & Adam\\
& regularization & L2\\
& regularization term & 0.0001\\
& learning rate & 0.001\\
& mini-batch size & 200\\
& maximum iterations & 10000\\
\midrule
\multirow{6}{*}{LightGBM}
& learning rate & 0.1\\
& number of leaves & 31\\
& feature fraction & 0.6\\
& L1 regularization term & 1.0\\
& L2 regularization term & 1.0\\
& minimum data in leaf & 20\\
\midrule
\multirow{5}{*}{\textbf{Logistic Regression (meta-classifier)}} & optimization & liblinear\\
& regularization & L2\\
& penalty C & 1.0\\
& tolerance & 0.0001\\
& maximum iterations & 100\\
\bottomrule
\end{tabular}
\caption{Meta and zero-level hyper-parameters for the classifiers in Figure~\ref{fig:stacking_train}.}\label{table_stacking}
\end{table}

\section{Style Breach Detection}

In this section, we describe our approach to the more complex task of finding the positions where style breach occurs, which we address using the supervised model from the previous Section~\ref{style_change_detection}. 

\subsection{Data}

We use the dataset from the PAN-2017 competition\footnote{\url{http://pan.webis.de/clef17/pan17-web/author-identification.html}} \cite{stein:2017m}, which consists of 187 documents each containing 1,000--2,400 word tokens. About 20\% of the texts have no style changes and the rest have between 1 and 8 changes. Switches of authorships\footnote{In this dataset, style change also means switch of authorship.} may only occur at the end of sentences, not within. The exact positions of the style changes in the multi-authored documents are provided as part of the dataset, but we did not use them for training.

This dataset is hard due to its small size and to class imbalance. Applying our model from the previous section on it poses further challenges as we have originally developed the model to identify the presence of changes in shorter texts (300--1,000 tokens) and with fewer style breaches (up to~3).

\subsection{Method}

Given a document to analyze, we first apply our model from the previous Section~\ref{style_change_detection} to predict whether there is style change in it. If style change is detected, we split the document in two: each half containing the same number of sentences. Then, we perform the same check for style change on each of the two parts, and if the results for both parts are negative then the exact position of the breach would be where the text was split in half. We repeat this procedure of splitting and searching for changes recursively until the length of the text fragment becomes less than 20 sentences, in which case, we return the middle point and we perform no more checks on the respective fragment.

Note that at training time, the model checks for the presence of changes on the full text, while at testing time it is applied to fragments of various sizes: starting with documents that are larger than those seen in training and going down to fragments whose size decreases exponentially. This discrepancy in the fragment sizes at training and testing makes the model's task harder. In order to alleviate the problem, we chose a relatively large minimum size for the text fragments of 20 sentences, assuming that shorter texts would not be easily handled by the model; we later confirmed this suspicion experimentally.

The next issue is that due to class imbalance, the model is much more likely to predict the positive class, which results in a lot of false positive, i.e., many non-existent style breaches being predicted in a document. On many occasions, the recursive procedure predicted an unreasonable number of breaches in a single document in the range of 20 or even 30, considering that the maximum number of style changes in a document in the training data was only 8. This is not completely unexpected though, as during training the model was never told the exact number of changes, just that there should exist at least one. Our strategy to cope with this was to increase the threshold for predicting that there is a style breach from 50\% to 75\%. This resulted in a significant drop in the average number of breaches predicted by the model from over 10 to 3.265.


\subsection{Results}

We evaluated the performance of our model for predicting the location of the style breaches using the following two evaluation measures:

\begin{itemize}
	\item \emph{WindowDiff} \cite{Pevzner:2002:CIE:636735.636737}, which is standard in general text segmentation evaluation, and returns an error rate between 0 and 1 (0 indicating perfect prediction) for predicting the exact location of the breaches by penalizing near-misses less severely compared to other/complete misses or to predicting more style breach locations than there are to be found. 
    \item \emph{WinPR} \cite{scaiano2012getting}, which computes common information retrieval measures, precision (WinP) and recall (WinR), and thus makes a more detailed, qualitative statement about the model performance.
\end{itemize}

\begin{table}[tbh]
\centering
\setlength{\tabcolsep}{10pt} 
\renewcommand{\arraystretch}{1.2} 
\begin{tabular}{lllll}
\toprule
&\textbf{windowDiff} &  \textbf{winP} & \textbf{winR} & \textbf{winF} \\
\midrule
BASELINE-rnd& 0.6088& 0.2779  & 0.5477 & 0.2366\\
BASELINE-eq&0.6345 & 0.3326  & \textbf{0.6368} & 0.2907\\
Stacking&\textbf{0.5719} & \textbf{0.3395}  & 0.6132 & \textbf{0.3302}\\
\bottomrule
\end{tabular}
\caption{Style Breach Detection: results for predicting the location of the breach.}\label{table_results}
\end{table}

\noindent We assessed our results by comparing them to the two baselines from \cite{stein:2017m}:
\begin{enumerate}
  \item BASELINE-rnd randomly places between 0 and 10 borders at arbitrary positions
inside a document.
  \item BASELINE-eq also decides on a random basis how many
borders should be placed (again 0--10), but then places these borders uniformly, i.e.,~so that all resulting segments are of equal size with respect to the tokens contained.
\end{enumerate}

Table~\ref{table_results} shows the average results we achieved by applying our model using 5-fold cross-validation as well as the scores for the two baselines above. 
We can see that our stacking approach managed to outperform the two baselines on both evaluation measures. Our results are also close to the ones achieved at the PAN-2017 edition \cite{karas:2017}, although they cannot be compared directly, as the systems that participated in the PAN competition were evaluated on a different test dataset.



\section{Conclusion and Future Work}

Detecting style change in texts is a difficult task for humans: we, ourselves, found it hard to discern between texts with and without style change while exploring the dataset. Nonetheless, our experiments have shown that it is possible for machine learning algorithms to achieve good performance for this problem.

The idea of applying a model recursively to find the exact style breach positions came to us as a natural experiment after tackling the simplified binary task of style change detection, for which we achieved an accuracy of 89\%. Our results for the more complex style breach position task are very close to the current state-of-the-art without the need for much adaptation of the original solution.


We believe that the results can be improved further if training is done on text pieces of different lengths, given that during validation, the recursive procedure has to be applied to smaller and smaller fragments of the original document. Tuning the model to work better for the case of imbalanced classes can also be a source of improvement. Another aspect we believe to be an important part of understanding the problem, and is yet to be explored, is the analysis and engineering of features based on the disparity between author takes on particular linguistic structures.

\subsubsection*{Acknowledgements.}
This work was supported by the Bulgarian National Scientific Fund within the project no. DN 12/9, and by the Scientific Fund of the Sofia University within project no. 80-10-162/25.04.2018.

\bibliographystyle{splncs04}
\bibliography{bibliography}






\end{document}